# Gabor-like Image Filtering using a Neural Microcircuit

C. Mayr[1,2], A. Heittmann[1], R. Schüffny[2]

1) Infineon Technologies AG, Corporate Research Munich

2) University of Technology Dresden
   Endowed Chair for Neural Circuits and Parallel VLSI-Systems

Corresponding author:    Christian Mayr
                         Dresden University of Technology
                         Department of Electrical Engineering
                         Circuits and Systems Laboratory (CSEL)
                         Endowed Chair for Neural Circuits and Parallel VLSI-Systems
                         Helmholtzstraße 10
                         01062 Dresden
                         Telephone number: +49 351 463 34943
                         Fax: +49 351 463 37260
                         Email: mayr@iee.et.tu-dresden.de

**Abstract: We present an implementation of a neural microcircuit for image processing employing Hebbian-adaptive learning. The neuronal circuit utilizes only excitatory synapses to correlate action potentials, extracting the uncorrelated ones, which contain significant image information. This circuit is capable of approximating Gabor-like image filtering and other image processing functions.**

## I. INTRODUCTION

Gabor wavelet decomposition has long been established as an efficient way of image compression and analysis. Experimental biological evidence also points to Nature's use of very similar convolution masks [1]. Technical implementations of neural circuits aim to copy some of the image segmentation/analysis/decomposition properties exhibited by mammalian visual pathways. VLSI-based spiking neurons so far have mainly exploited the synchronization patterns of locally coupled spiking neurons using various synapse adaptation rules [2,3]. However, this type of network pattern is relegated to simple image analysis. More complicated neurons and synapses, as well as more complex network topologies are needed to achieve high-level, non-trivial image processing functions [4]. In Section II, a neural microcircuit capable of extracting the uncorrelated pulses from two pulse streams [5] and its modification for a hardware implementation are briefly described. Its application to image filtering tasks up to the complexity of Gabor masks and the relevant processing cascade is given in section III. Also, several simulation results are presented, underlining the efficacy of the proposed pulsed image processing for fast, low resolution, as well as slow, high resolution image analysis functions.

## II. THE MICROCIRCUIT

A. Neuronal Circuit and Adaptation Rules

Recurrent, stereotyped neural microcircuits occurring in biological neural networks have been described in [6]. One important aspect of these microcircuits is their individual simplicity, contrasting with the complex processing functions a network of these circuits is capable of. The ones analyzed in [6] are using excitatory and inhibitory synapses, but purely inhibitory microcircuits have also been found in mammals [7]. In both references, [6] and [8], it is postulated that extracting correlations between inputs is one of the major processing functions, with [8] presuming that inhibitory synapses are necessary to achieve the synchronous activity among neurons required for correlation detection [2,4]. This is in contrast with the work reported in [2,3,5], where purely excitatory synapses also achieve synchronous neuronal activity for correlated inputs. So, while neural microcircuits consisting solely of excitatory synapses have not yet been found, there are no theoretical obstacles to their implementation. Additionally, excitatory synapses arranged in a feed-forward manner as contained in the microcircuit described herein make for rapid information processing [9]. So, to explore the complex processing possible through networks of microcircuits, and to take advantage of feed-forward excitatory neural structures, the following microcircuit (introduced in [5]) has been implemented:

Fig. 1.

The neural microcircuit consists of simple non-leaky Integrate-and-Fire (IAF) neurons connected by two types of adaptive synapses (Fig.1). The dynamics of the IAF neurons are given by multiplying the pulses running along the synapses with the respective synapse weights and integrating them on the neuron membrane. Once the membrane potential reaches a firing threshold θ of one, the neuron emits an output pulse, immediately resets the integrator and is open for new inputs (no refractoriness period, i.e. temporal blocking of the membrane integrator).

The adaptation rule of the first two synapses ($W_{31}$ and $W_{32}$), here called a membrane adaptation, is given in equation (1) (Indices are shown for the synapse connecting neurons 1 and 3, expressed by $W_{31}$):

$$\tfrac{d}{dt}W_{31} = -\gamma \cdot W_{31} + \mu \cdot (a_3 - \tfrac{\theta}{2}) \cdot \chi(X_1) \tag{1}$$

It is a basic Hebbian learning rule intended to synchronize pulses with almost constant phase relationships [3], with γ as decay term, μ as learning rate, $a_3$ denoting the membrane accumulator state, θ the positive/negative learning threshold, and χ being the indicator function of neuron 1 output $X_1$, one if neuron 1 exhibits a pulse, zero otherwise. The decay term 'forgets' the learned weight if it is not reinforced by pulse activity, while the second term acts as correlator between the accumulator state of neuron 3 and the pulses exhibited by neuron 1. The accumulator of neuron 3 has to be high (i.e. close to its firing threshold) when neuron 1 fires, for the weight W31 to increase, which reflects the Hebbian aim of increasing a synaptic weight if the presynaptic neuron takes part in firing the postsynaptic one.

In this particular application, (1) is employed to extract correlated pulses from the output pulse streams of neurons 1 and 2. To illustrate the correlation function of neurons 1 through 3 governed by (1), let's assume neuron 3 has just emitted a pulse and has a membrane potential $a_3$ close to zero. If neuron 2 emits a pulse next, the corresponding weight $W_{32}$ is increased (μ<0 and $a_3$-θ/2<0) and $a_3$ pushed above θ/2. If neuron 1 emits a pulse next, its corresponding weight is also increased (μ>0 and $a_3$-θ/2>0) and neuron 3 is pushed above the firing threshold. Only this particular phase relationship (i.e. a pulse of neuron 2 followed by a pulse of neuron 1) results in neuron 3 emitting pulses, thus neuron 3 emits only pulses correlated between neurons 1 and 2. This adaptation rule, when applied to image processing, acts as an inverted first-order difference operator, i.e. the output activity of neuron 3 decreases with the steepness of the gradient between neurons 1 and 2, an effect which is used in [10] for a somewhat similar, pulse-timing based rule to detect edges and establish image depth from pulse correlations.

The second adaptation rule acting on synapse $W_{41}$, the dendrite adaptation, is given as:

$$\tfrac{d}{dt}W_{41} = -\gamma \cdot (W_{41} - W_\infty) + \mu \cdot (X_3 \cdot W_{43} + X_2 \cdot W_{42} - I_\theta) \cdot W_{41} \cdot \chi(X_1) \tag{2}$$

which works in such a way that, with no pulses present, the first term draws $W_{41}$ asymptotically to $W_\infty$, letting pulses from neuron 1 pass to neuron 4. If, however, the second term is added through a pulse event $\chi(X_1)$, with the threshold current $I_\theta$ equal to 0.02, i.e. less than $X_3 \bullet W_{43}$ or $X_2 \bullet W_{42}$, the weight $W_{41}$ is decreased. This means, that if a pulse is detected further up the dendritic tree ($W_{43}$ or $W_{42}$), this pulse blocks any that would be transmitted by $W_{41}$ to neuron 4. Compared to the membrane adaptation, the dendritic adaptation acts very fast, as is evident by the different adaptation parameters µ given in Fig. 1. This type of adaptation operates on single pulses, producing a quasi-digital gating function [4, section 19.3.2]. $W_{42}$ is entered into the circuit to precharge the dendritic adaptation, also mitigating the delay inherent in propagating the correlated pulse across neuron 3.

Further, we define the correlation between the pulses of neurons 1 and 3 as follows:

$$C_{13} = \tfrac{1}{T_1 - T_0} \int_{T_0}^{T_1} \chi(X_1(\tau)) \cdot \chi(X_3(\tau)) d\tau \tag{3}$$

With the normalized correlation given as $\overline{C_{13}}=C_{13}/C_{11}$, and the normalized decorrelation defined as $\overline{D_{13}}=1-\overline{C_{13}}$, the net effect of theses rules and synapses is given in Fig. 2, i.e. if there are uncorrelated pulses from neuron 1, these are transmitted across the neural microcircuit, uncorrelated (super numerous) pulses from neuron 2 are ignored. The microcircuit acts as a pulse subtractor, restricted to the positive domain, i.e. once neuron 2 exhibits more activity than neuron 1, the output neuron 4 simply stays at its lowest activity level. For a more detailed discussion of the neural microcircuit, please see [5].

Fig. 2.

B. Digital Implementation and Input Circuitry

One of the adaptation rules, the membrane adaptation, has been implemented in analog hardware for a different application, exhibiting its veracity compared to the simulation [3]. However, for the Integrated Circuit (IC) implementation of the microcircuit, due to size and time constraints, a pulsed pseudo-digital representation of the microcircuit has been carried out, exhibiting the same behaviour. This microcircuit is part of a neural processing unit (NPU, Figure 3), with additional functions for pulse weighting and a digitally realized neuron, both carried out by the same digital accumulator, which either transmits the weighted signal (synapse function) or a single pulse if the accumulator reaches a certain threshold (neuron function).

Fig. 3.

The weighting can be governed by external configuration signals, i.e. further adaptation rules carried out in digital computation in an external FPGA. Also, a simple pulsing pixel cell is part of the NPU, consisting of a CMOS photo sensor, accumulator, and thresholded pulse generator [11]. This pixel cell provides the pulsed input for the neural processing elements, i.e. the neural microcircuit and the weighting/digital neuron. The pulse frequency of the pixel cell output is linearly dependent on the photo current produced by the CMOS photo sensor.

III. Image Filtering

A. Pulsed Edge Detector

A group of the neural microcircuits can be used to construct a simple edge detector by connecting their inputs to adjacent pixels, stacking them according to Fig. 4a, feeding their respective input neurons 1 (+) and 2 (-) with pulsed representations of the greyscale image and summing their outputs. These pulsed greyscale images are obtained by supplying an AHDL representation of the pulsing pixel cell [11] with input current linearly based on pixel brightness (greyscale value). Fig. 4b shows the output of the edge detector if an edge moving from left to right is presented to it.

The topological linking of the pixel cells to the microcircuits and (in case of higher-level processing) the linking of microcircuits among themselves is carried out via an Address-Event-Representation (AER) of their respective output pulses, with pulses distributed across the microcircuits (to single or multiple destinations) in a packet-based way, similar to the AER described in [12].

Fig. 4.

As can be seen, the pulse response rises when the edge enters the neighborhood of the detector, with a defined maximum (the 'measured' curve, which denotes a pulsed simulation as described below). The ideal curve is obtained by feeding the same moving edge to a continuous representation of the microcircuit, and computing the resultant pulse counts. The slopes are due to the exact modelling of the pixel cell, which includes metal stack effects of modern CMOS technologies. This means, that as the photo diode is contained in the substrate, i.e. at the bottom of the IC, the light reaching it will not only be the light entering directly, but also indirect light reflected off the metal layers at the edges of the pit the diode is contained in. For practical purposes, this can be thought of as a Gaussian smoothing with a σ of about 1.4 pixels, resulting in slopes in the response curve, rather than a sharp transition as the edge in the input image enters the edge detector. The discrepancy between ideal and pulsed, i.e. 'measured', curve is also due to this random light scattering effect, which, especially at very high and low current levels, leads to noise on the photo current and thus to jittering of the pulse times. This upsets the pulse order required for correct operation of the microcircuit as expressed by (1) and (2), thereby leading to false pulse responses for low brightness contrast (at the border of the edge) and pulses lost for high brightness contrast, as compared to the ideal curve.

A note on the method by which these results were obtained: An implementation of the microcircuit-IC has been carried out in a 130nm CMOS technology to tape-out level, but cancelled due to funding constraints. The results presented herein are based on system-level Mentor ModelSim simulations of VHDL/AHDL code of the IC. Major components of the IC, however, have been implemented and verified previously, like the Adress-Event-Representation employed for pulse communication [3], or the pulsing pixel cell and the adapting synapses as mentioned above. Where applicable, the VHDL/AHDL code has been augmented by measurement results to enhance its veracity.

The effectiveness of this edge detector can be enhanced by feeding both the summed outputs of a positive (analog to Fig.4) and negative edge detector to one of the neural microcircuits (Fig. 5), which pools them. Both edge detectors are located at the same spatial coordinates, i.e. on top of each other, having access to the same pixel pulse trains, the separation between both in Fig. 5 only denotes the processing scheme. This way, the negative edge detector will respond to the occasions when the pulse order is jittered for the positive one, as explained above. The 'pooling' microcircuit then subtracts the adjustment signal provided by the negative edge detector from the original answer of the positive edge detector, resulting in a better translatory precision, because the jittered false pulse responses cancel each other. However, this method also reduces the maximum response, where pulses are already lost due to jittering, and still more are subtracted.

Fig. 5.

B. Pulsed Processing Pyramid for Gabor Filtering

By setting up a processing pyramid, i.e. several ordered processing steps, of these microcircuits and introducing pathways to the neurons in the extended neighborhood, more complex image filtering functions can be accomplished, in particular, Gabor wavelet decomposition of the image can be realized as a pulsed image computation. This is somewhat similar to the image processing described in [13], where select synapses, each connected to a single pixel realizing a retina function, are linked together to produce receptive fields in differing orientations. The discussed implementation differs in three aspects from [13], first, the receptive fields are not fixed in hardware, they can be arbitrarily chosen with the configurable AER as discussed above. Second, both the input pixel cells and the microcircuits are realized on the same IC, so the receptive fields can be built on this single IC. Third, the communication between pixels and higher-level processing is pulse-based, quasi-digital, not analog.

In the case of Gabor filtering, the one-sidedness of the pulse subtraction becomes a more serious issue, so that using opposing masks at the same spatial coordinates is not simply advisable for improving noise errors, as in the edge detectors above, but rather to counter an inherent, systematic flaw of this method of image convolution. Consider for example, a 1D convolution with mask (1 –2 1), which could be a slice through a simple even Gabor filter [1] or which could be added in the form (1 -2 1 0 0)+ (0 0 1 -2 1)=(1 -2 2 -2 1) to produce more complex Gabor filters. This mask can be realized with a pair of NPUs in the form (+ -- +), where both negative inputs (-) access the same pixel. If an input pattern of (3 2 1), respectively its pulsed representation, were presented to the ideal mask, the response would of course be 0. Using the neural microcircuits, however, the first NPU would deliver 3-2=1, whereas the result for the second NPU is 1-2=0, because of the one-sidedness of its subtraction (Fig. 2), so the summed response of the NPU mask would be 1. If we introduce a second, negative mask (- ++ -), its result would be 1 as well, so we get the correct result of 0 by subtracting the result for the negative mask (- ++ -) from the positive mask response. This procedure will not alter the response for a perfect fit to the positive mask, since the negative mask responds with 0 in such a case. This correction is not perfect, an input pattern of (2 2 1) will result in –1 for the ideal mask, whereas even the corrected signal (pos-neg) for the NPUs is 0. The method described above for setting up an 1D filter using the microcircuits can, of course, be extended to the two-dimensional case.

Since Gabor amplitude response is of most use in image analysis, the next step would be computing the absolute value Gabor response $R_{abs}$ from the positive and negative Gabor mask responses $R_+$ and $R_-$. This absolute value is of course defined as:

$$|R_+ - R_-| = \begin{cases} R_+ - R_- & \text{for } R_+ \geq R_- \\ R_- - R_+ & \text{for } R_+ < R_- \end{cases} \quad (4)$$

If we input R+ and R - once in every direction to a neural microcircuit and sum the results, the one-sided subtraction results in the same computation, which incidentally also corrects the positive and negative mask responses by subtraction of the opposing mask, as described above:

$$(R_+ - R_-) + (R_- - R_+) = \begin{cases} R_+ - R_- + 0 & \text{for } R_+ \geq R_- \\ 0 + R_- - R_+ & \text{for } R_+ < R_- \end{cases} \quad (5)$$

In Figure 6, the actual implementation of the processing pyramid is depicted, realized on the neural processing IC consisting of 128*128 NPUs. The figure shows a greyscale representation of a pulse histogram of the NPU output pulses across the IC. A pulse histogram, i.e. a location-specific accumulation of pulses is a convenient way to represent the processing carried out in the various microcircuits, similar to the RC-reconstruction of an AER code as outlined in [12]. Sections of the NPU matrix and their attendant pulse routing have been configured to perform the processing steps described above. These processing steps consist of (clockwise from top left) the pulsed image input, generation of the image convolution response to the regular submasks of the Gabor mask (top right, submask principles as mentioned above), summation of those submask answers in the positive and negative Gabor mask (bottom left), subtraction of positive and negative mask in both directions (bottom right, see also (5)), and summation of these revised positive and negative masks into the final Gabor convolution answer (middle left, below original image).

Fig. 6.

## C. Examples for Pulsed Gabor Filtering

To show the accuracy of this pulsed convolution method as compared to conventional image convolution, Figure 7 gives a comparison between original image, amplitude of the Gabor response for conventional convolution, and the pulsed Gabor answer:

Fig. 7.

The pulsed computation of the Gabor mask exhibits two distinct advantages compared to a conventional image convolution, namely robustness to temporal noise (compare Fig 8b, this convolution response has been computed with a 20% white noise level on the pixel currents, but shows little difference to the result from Fig. 7c, due to the noise cancellation carried out via positive and negative masks). Also, the pulsed computation inherently operates at different levels of resolution. As can be seen from the picture shown in Fig. 8a (taken after the brightest pixel in the original image had emitted three pulses), the most distinctive features evident in the full Gabor response are also evident in this early response, while the more detailed Gabor features do not show yet. So this pulsed computation makes for rapid, low resolution, and slow, high resolution image processing, which is analogous to biological neural nets [9,6].

Fig. 8.

## IV. CONCLUSION

We have presented a pulse-processing neural microcircuit as part of a neural processing IC. The neural microcircuit is based on two information processing principles postulated from biological evidence, Hebbian pulse correlation [4, section 13.5.1] and dendritic pulse gating [4, section 19.3.2]. While performing a simple pulse correlation/decorrelation individually, suitably shaped networks of these microcircuits are capable of realizing complex image filtering tasks such as Gabor wavelet decomposition as pulse-based computation. These networks employ some of the principles postulated from biological evidence, e.g. the building of complex masks through several simpler interim steps in a layered configuration, and the use of directly opposing masks at the same image location (similar to On/Off centers [1,8,9]) to cover the whole spectrum of convolution mask responses to a given image. While making no claim as to the veracity of the chosen approach with respect to the mammalian visual pathways, the developed Gabor operator exhibits some of the characteristics of said pathways, namely robustness, parallelism, and fast, low resolution, as well as slow, high resolution image convolution.

Fig. 1. The neural microcircuit.

Fig. 2. Results of neural microcircuit, normalized correlation and decorrelation between neurons 1 and 3, as well as the corresponding input pulse rates for neurons 1 and 2 and output neuron 4.

Fig. 3. Schematic of Neural Processing Unit.

Fig. 4. Edge detector and translatory response dependent on the location of the presented edge (physical location of edge detector corresponds to 10 pixels displacement)

Fig. 5. Positive and negative edge detector and resulting improved translatory response.

Fig. 6. VHDL simulation of pulsed Gabor convolution on the IC, with the numbers indicating Pixel/NPU location.

Fig. 7. Comparison of original image (a), amplitude response of an ideal Gabor image filter (b), and pulsed realization using the neural microcircuit (c).

Fig. 8. Results for computing pulsed Gabor filter answers with very little input pulses (a), and for computing it with input pulses corrupted by noise (b).

Fig. 1)

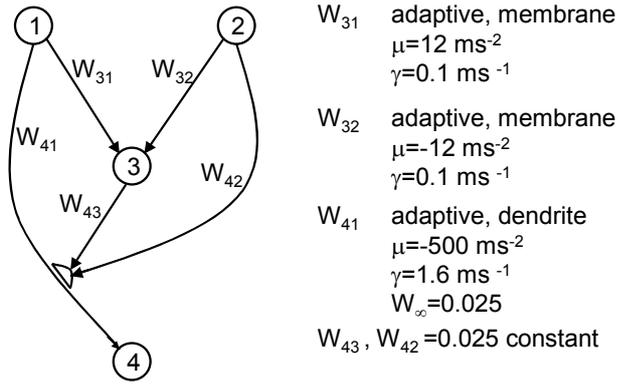

W$_{31}$  adaptive, membrane
  $\mu$=12 ms$^{-2}$
  $\gamma$=0.1 ms$^{-1}$

W$_{32}$  adaptive, membrane
  $\mu$=-12 ms$^{-2}$
  $\gamma$=0.1 ms$^{-1}$

W$_{41}$  adaptive, dendrite
  $\mu$=-500 ms$^{-2}$
  $\gamma$=1.6 ms$^{-1}$
  W$_\infty$=0.025

W$_{43}$, W$_{42}$ =0.025 constant

Fig. 2)

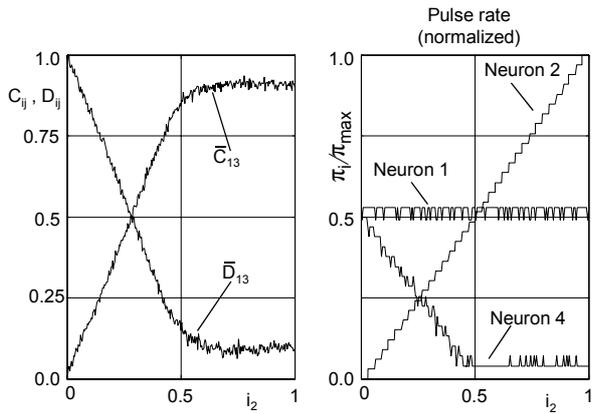

Fig. 3)

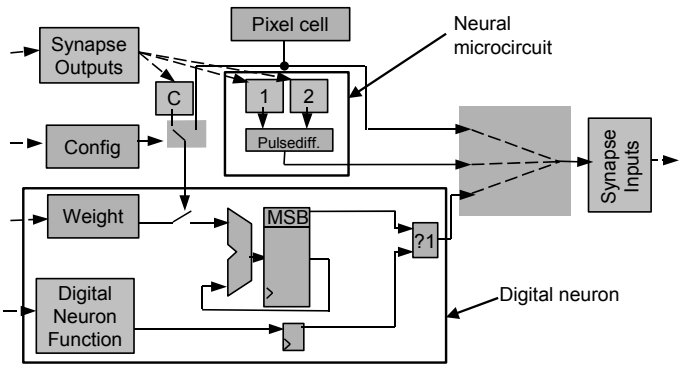

Fig. 4)

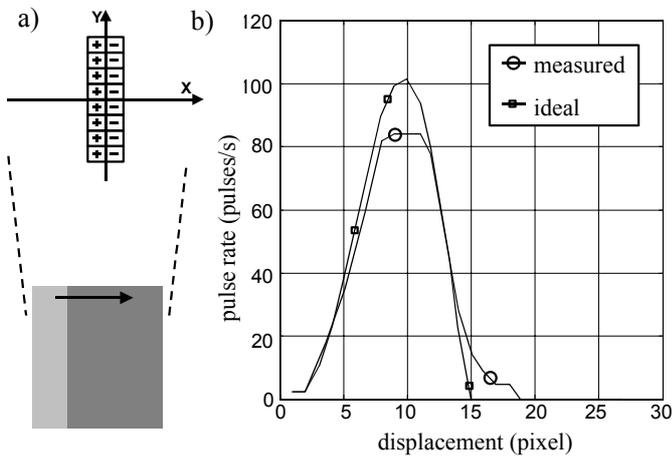

Fig. 5)

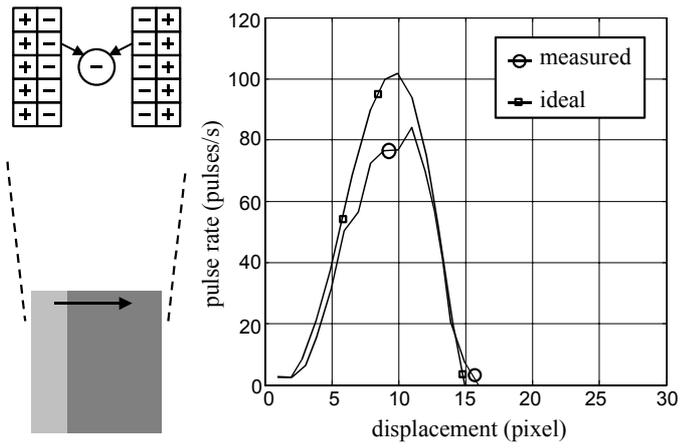

Fig. 6)

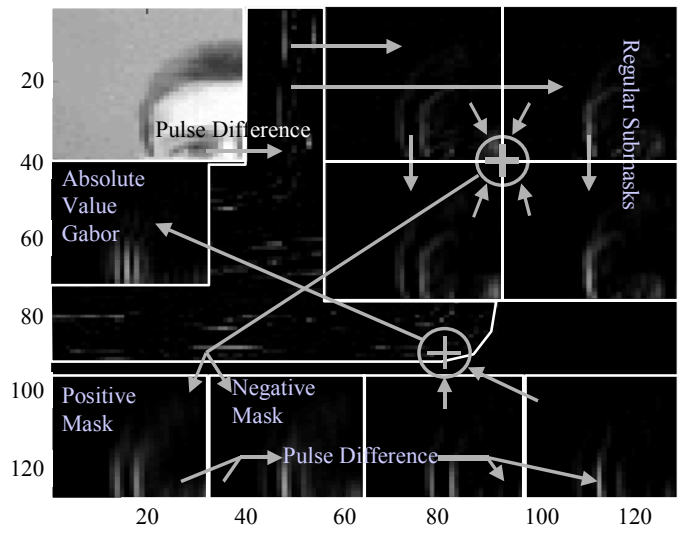

Fig. 7)

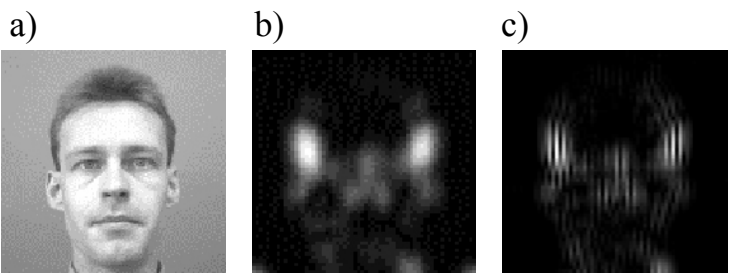

a)      b)      c)

Fig. 8)

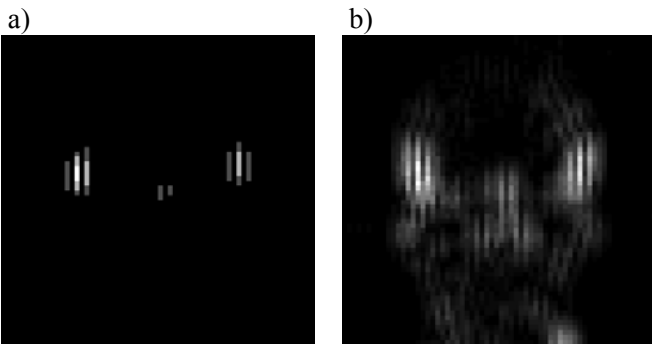

a)      b)